\documentclass[10pt,twocolumn,letterpaper]{article}

\usepackage{cvpr}

\usepackage[utf8]{inputenc}
\usepackage[T1]{fontenc}
\usepackage{graphicx}
\usepackage{url}
\usepackage{booktabs}
\usepackage{amsmath}
\usepackage{amsfonts}
\usepackage{nicefrac}
\usepackage{microtype}
\usepackage{xcolor}
\usepackage{pgfplots}
\pgfplotsset{compat=1.18}

\makeatletter
\providecommand{\@LN@col}[1]{}
\providecommand{\@LN}[2]{}
\makeatother

\definecolor{cvprblue}{rgb}{0.21,0.49,0.74}
\usepackage[pagebackref,breaklinks,colorlinks,citecolor=cvprblue]{hyperref}
\usepackage[capitalize]{cleveref}

\def\paperID{26}
\def\confName{3DV\xspace}
\def\confYear{2027\xspace}

\title{Everybody Tracking Every Body}

\author{
Daeyun Shin\textsuperscript{1} \quad
Yunhan Zhao\textsuperscript{1} \quad
Shu Kong\textsuperscript{2} \quad
Alexander C. Berg\textsuperscript{1} \quad
Charless Fowlkes\textsuperscript{1}\\[2pt]
\textsuperscript{1}University of California, Irvine \qquad
\textsuperscript{2}University of Macau
}

\begin{document}

\maketitle

\begin{abstract}

We address the problem of 3D body pose estimation of multiple interacting people from their egocentric views with centralized coordination.
Each individual wears a camera recording egocentric video and IMU data. Processing this video with VIO SLAM provides high-quality tracking of each egocentric camera through space. 
The first-person view from one individual provides third-person observations of other people,
although these exocentric observations are sparse, intermittent, and of highly variable reliability as both cameras and subjects move. To integrate these synchronized data streams, we propose a diffusion-based approach that fuses estimates of pose based on head motion derived from egocentric camera motion with exocentric pose observations, conditioning on both observation content and reliability. Our model is trained on a mixture of single-person motion-capture data and multi-person video in order to learn rich priors for body motion trajectories and video observation reliability. Evaluation on challenging multi-person datasets suggests our fusion approach improves over motion-only and vision-only baselines in terms of both absolute and relative pose accuracy.

\end{abstract}

\section{Introduction}
\label{sec:intro}

\begin{figure*}[tp]
\centering
\includegraphics[width=0.95\textwidth]{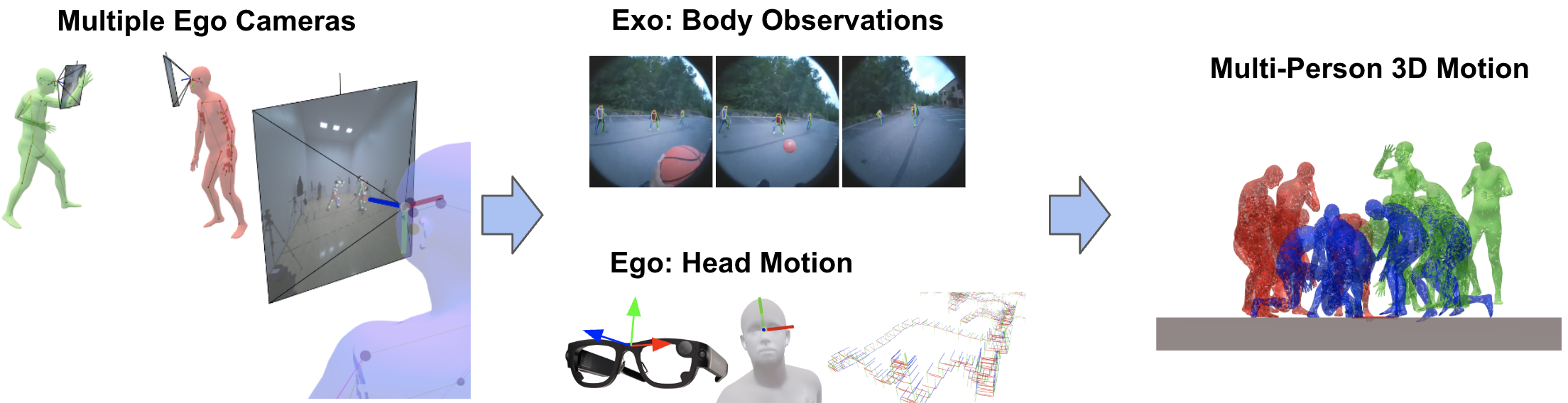}
\vspace{-2mm}
\caption{\small
\textbf{Problem overview.}
In a multi-person egocentric setting, each person's head-mounted camera provides third-person observations of others, while their own body motion is inferred from head trajectory alone. Can we fuse these complementary signals to enable global scene understanding and recover full-body 3D motion for all participants?}
\label{fig:overview}
\end{figure*}

Humans navigate the physical world from visual and proprioceptive sensory data perceived egocentrically. From this perceptual data, we self-orient to estimate our position in allocentric coordinates with respect to other people and the environment around us. This ability to reason about state in ``world coordinates'' is valuable in ensuring representations are persistent over periods of time when people or parts of the scene are occluded or outside our egocentric field-of-view. 

This setting also arises in understanding human activities on wearable camera platforms such as smart glasses, which are making egocentric video increasingly common in collaborative environments. When multiple people in a shared environment all wear a camera, a natural structure emerges: each person's first-person view provides third-person observations of others (\cref{fig:overview}), while their own body falls almost entirely outside their field of view. The same person is simultaneously an unobserved subject from their own perspective and an observed target from others'. This ego--exo duality creates complementary signals for body estimation that few existing methods exploit jointly.

Existing methods have explored full-body pose estimation from egocentric or exocentric observations alone. A prior learned from motion capture data (such as EgoAllo~\cite{egoallo}) can be used to estimate the wearer's body pose from their head-mounted camera trajectory, using the statistical relationship between head motion and body movement. This provides an estimate at every timestep with good absolute positioning (the body is attached to the known camera location) but limited accuracy in relative body pose, since head motion constrains limb positions only loosely. Alternatively, third-person trackers (such as CoMotion~\cite{comotion} and SAM 3D Body~\cite{yang2026sam3dbody}) can detect and estimate 3D body pose for people visible in each camera's video stream, a scene-centered signal. When a person is visible, this provides accurate relative body pose, but coverage is incomplete (people move in and out of each other's field of view) and absolute positioning is often limited by resolution/distance and occlusion due to the observer's viewpoint.

Neither source alone is sufficient. The head-motion prior has full temporal coverage but limited pose accuracy. Visual observations of the body have good pose accuracy but sparse temporal coverage (approximately 24\% of person-frames have no observation from any camera in our evaluation setting). Here we tackle the open problem of combining egocentric data from multiple people (everybody) to estimate the body pose trajectories for all the individuals simultaneously (every body).

We develop a conditional diffusion model that integrates an egocentric camera trajectory with a variable number of noisy exocentric observations to estimate full body pose. Our model is trained primarily on single-person motion capture data (AMASS~\cite{amass}) to learn the distribution of human motion and on a smaller quantity of multi-person egocentric video footage (EgoHumans~\cite{egohumans} and Harmony4D~\cite{harmony4d}). This allows the model to learn rich priors for full-body motion trajectories and calibrate the reliability of sparse, noisy exo-video observations. We also evaluate alternative conditioning baselines which use direct imputation or diffusion posterior sampling. Experimental evaluation shows that our proposed conditional model is effective at synergistically fusing both sources of information and outperforms approaches that only address individual aspects of the problem.

To summarize our contributions: (1) We present a novel problem formulation which utilizes multi-camera egocentric data to jointly track multiple people, including individuals that are not visible in any view. (2) We introduce a conditional diffusion architecture that integrates variable numbers of egocentric and exocentric observations across multiple dynamic cameras to predict consistent multi-person 3D motion in a global coordinate frame. (3) We demonstrate the empirical benefit of fusion on improved estimation accuracy, outperforming strong baselines built on conditional imputation.

\section{Related Work}
\label{sec:related}

\begin{figure*}[tp]
\centering
\hfill
\includegraphics[width=1.0\textwidth]{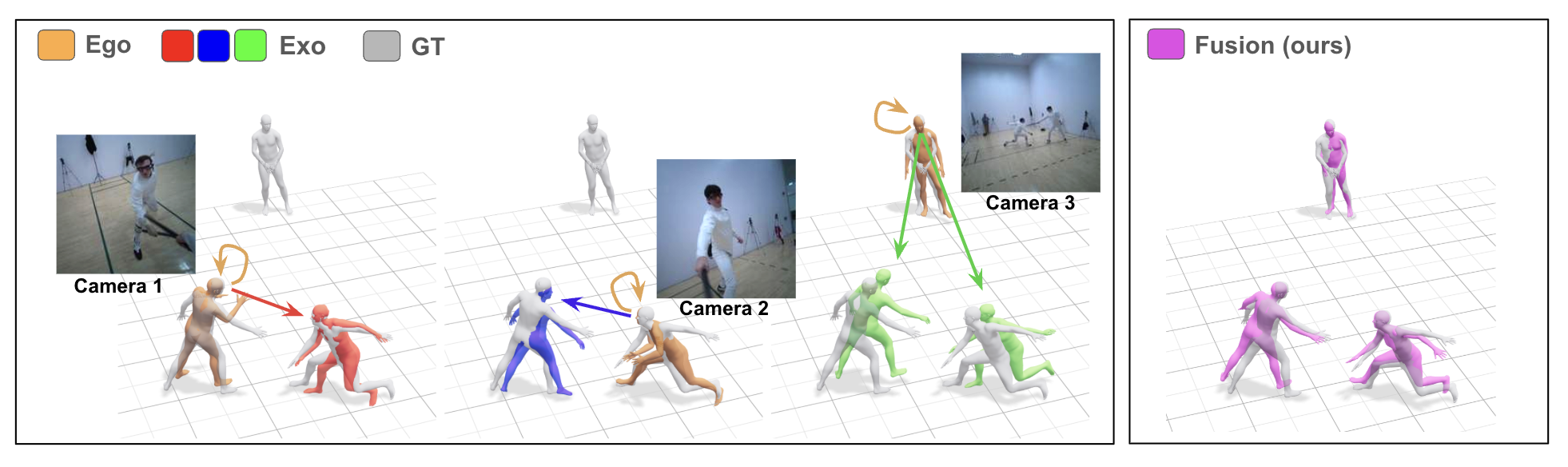}
\vspace{-7mm}
\caption{\small
Each camera provides information about the wearer's pose (ego) and the pose of others visible in the scene (exo).
Egocentric camera trajectories provide accurate head location but inaccurate limb poses (e.g., camera 2 ego estimate).  Exocentric detection provides good joint angle estimates but suffers from absolute position errors due to scale ambiguity and global offset (e.g., green observations) and missing predictions (out-of-view or occluded). In our experiments 24\% of person-frames are missing exo-predictions (e.g., wearer of camera 3).
Our method fuses ego and exo data into a single global estimate (purple), providing improved pose for both observed and out-of-view individuals.
}
\label{fig:comparison}
\end{figure*}

\noindent\textbf{Pose estimation with moving cameras.}
Estimating human motion from moving cameras requires recovering body motion and camera motion in a shared world frame. Early work fused body-worn IMUs with a moving monocular camera through joint graph optimization~\cite{vonMarcard2018}. Recent monocular methods recover globally coherent human motion by jointly estimating camera trajectories and human pose, including GLAMR~\cite{glamr}, SLAHMR~\cite{ye2023decoupling}, TRAM~\cite{wang2024tram}, GVHMR~\cite{shen2024gvhmr}, TRACE~\cite{Sun_2023_CVPR}, Human3R~\cite{chen2025human3r}, JOSH~\cite{liu2025josh}, and DuoMo~\cite{wang2026duomo}. These works address world-space reconstruction from dynamic cameras, but mostly assume an exocentric camera observing the people being reconstructed. Our setting differs because the cameras are worn by the subjects themselves: each person's camera trajectory anchors their own global motion, while other wearers' videos provide intermittent exocentric observations. This lets us estimate body motion even when a person is not visible in any camera view.

\noindent\textbf{Egocentric and exocentric body estimation.}
Egocentric methods estimate the camera wearer's body pose, which is often outside the field of view. Prior work uses social interaction cues~\cite{ng_you2me_2020}, egocentric benchmarks such as EgoBody~\cite{Zhang:ECCV:2022}, sparse head and hand tracking as in EgoPoser~\cite{egoposer}, or head-trajectory-conditioned priors as in EgoAllo~\cite{egoallo}. These methods provide dense estimates for the wearer but are weakly constrained in detailed limb pose. In contrast, exocentric methods such as HMR~\cite{hmr}, VIBE~\cite{vibe}, CoMotion~\cite{comotion}, and SAM 3D Body~\cite{yang2026sam3dbody} estimate visible people from third-person images or videos, providing more direct body evidence but only when the subject is visible and localized. Our method combines these complementary signals: an egocentric prior gives full temporal coverage for each wearer, while exocentric trackers provide sparse, confidence-weighted pose observations of other people from the remaining cameras.

\noindent\textbf{Conditional body motion generation.}
Generative motion models offer a way to combine motion priors with partial observations. Diffusion-based models such as MDM~\cite{mdm} and PriorMDM~\cite{priormdm} learn strong human motion priors, while body-conditioned methods such as HumanMAC~\cite{Chen_2023_ICCV}, OmniControl~\cite{omnicontrol}, and CondMDI~\cite{condmdi} generate motion from observed frames, joint constraints, or in-betweening targets, often assuming sparse but accurate conditioning signals. Other methods handle noisier observations: HuMoR~\cite{Rempe_2021_ICCV} reconstructs plausible motion with a learned prior, RoHM~\cite{rohm} and Saadatnejad et al.~\cite{saadatnejad_diffusion_pose} use diffusion for noisy reconstruction or forecasting, GENMO~\cite{li2025genmo} treats estimation as constrained generation from signals such as video, keypoints, and 3D keyframes, and DuoMo~\cite{wang2026duomo} jointly models body and camera motion. Our setting differs because the conditioning signals are predicted poses from multiple moving egocentric cameras: they are noisy, intermittent, confidence-weighted, and missing whenever a person is outside all views. We train a conditional model that accounts for this by implicitly learning to accommodate noise in observations rather than treating them as hard constraints.

\section{Method}
\label{sec:method}

\cref{fig:comparison} visualizes the strengths and weaknesses of egocentric motion and exocentric video in estimating body motions. In multi-person scenarios with head-mounted cameras, one person's egocentric view provides close-range exocentric observations of others. Such perspectives naturally capture rich visual detail, but both cameras and bodies move, and limited field of view causes people to frequently appear and disappear from view. While visual observations are intermittent, ego-motion can always be relied upon, yet it doesn't provide the full picture required for precise reconstruction. We formulate fusion in probabilistic terms and describe our implementation based on conditional diffusion.

\subsection{Problem formulation and modeling}
We consider $N$ people moving in a shared environment, each wearing a head-mounted camera (Meta Aria glasses in our experiments, with $N$ ranging from 2 to 4). All camera feeds and pose estimates are synchronized and expressed in a common world coordinate frame ($z$-up, ground-plane at $z=0$). For each person $i$ at time $t$, we have access to an egocentric video stream $I^i_t$ and a camera pose trajectory $C^i_t$ estimated from visual-inertial odometry (VIO/SLAM).

The goal is to estimate the body state $S^i_t = (\theta^i_t, \beta^i_t)$ and root-joint pose $L^i_t = (R^i_t, t^i_t)$ for every person at every timestep. We represent the body using the SMPL-H model~\cite{smpl,smplh}, whose state at each time is a 496-dimensional vector describing body shape ($\beta^i_t$) and joint angles ($\theta^i_t$) along with contact flags and hand rotations.
We relate the egocentric camera pose to the body root-joint pose through a deterministic mapping $C^i_t = \Phi_{S^i_t}(L^i_t)$ and its inverse $L^i_t = \Phi^{-1}_{S^i_t}(C^i_t)$. This mapping is defined via the Central Pupil Frame (CPF), the midpoint between the pupil center vertices of the posed SMPL mesh, with the same rotation as the SMPL head joint. The CPF serves as the kinematic root of the body (instead of the standard SMPL pelvis root), so the body ``hangs'' from this head-anchored frame. 

We model each person's body state independently, conditioned on the observations from all cameras:
\[
P(S, L \mid I, C) = \prod_i P(S^i, L^i \mid I^1, \ldots, I^N, C^1, \ldots, C^N).
\]
We note this simplified independence assumption can't explicitly model inter-person interactions (e.g., constraining poses to not intersect). Instead we focus on the primary coupling between people arising through their mutual visibility: person $j$'s camera may observe person $i$, providing exocentric estimates $(\hat{S}^{ij}_t, \hat{L}^{ij}_t)$ when $i$ is visible in camera $j$ at time $t$.

When only the egocentric camera trajectory is available, we would like to estimate body motion using a conditional model $p_\theta(S^i | I^i, C^i)$. The predicted body is then globally positioned by kinematics from the head joint: $\hat{L}^i = \Phi^{-1}_{{S}^i}({C}^i)$ (analogous to the model proposed in EgoAllo \cite{egoallo}). The ego-camera trajectory provides full temporal coverage and accurate absolute positioning of the person, but there is significant uncertainty in the detailed body pose of the limbs.

When an individual is visible in another person's view, we can utilize standard 3D pose estimation and tracking techniques to estimate multiple body poses from exocentric RGB observations. In this setting, data association between body trajectories in video from a camera $C^j$ and individuals can be easily performed by assigning detections based on the distance between the CPF location estimated from the exo-detection and the camera trajectories of each individual ${C}^i_t$ relative to camera $C^j$.

\begin{figure*}[tp]
\centering
\includegraphics[width=\textwidth]{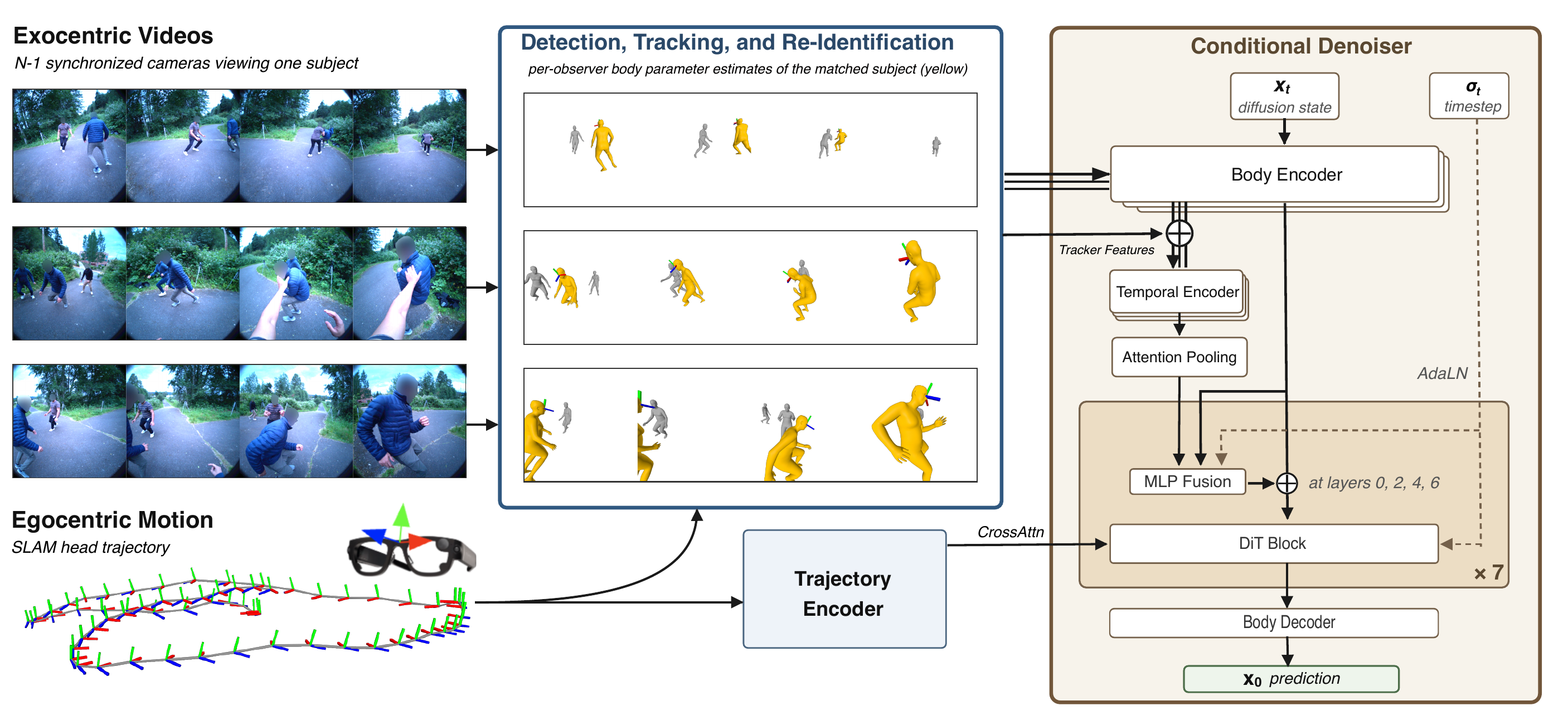}
\vspace{-7mm}
\caption{\small
\textbf{Model architecture.} Exocentric views from participants' cameras are processed by a 3D pose model (CoMotion) to produce per-frame body motion estimates, and these observations are associated with a given individual's egocentric motion. The sparse and variable number of estimates are encoded along with the ego trajectory and used to condition a DiT denoiser to estimate the individual's body trajectory.
}
\label{fig:architecture}
\end{figure*}

For person $i$ observed in camera $j$, a detector predicts body pose $\bar{S}^{ij}_t = f(I^j_t)$ in camera $j$'s coordinate frame. We can then model $P(S^i|I^j,C^i,C^j) \approx p_\theta (S^i|\hat{S}^{ij},\hat{L}^i)$  based on the estimates $\hat{S}^{ij} = T_{\mathbf{C}^{-1}_i \mathbf{C}_j} (\bar{S}^{ij})$ and $\hat{L}^i=\Phi^{-1}_{\hat{S}^i}(C^i)$ where $T$ maps the pose to person $i$'s root joint frame via the relative transformation between the two cameras.
This can produce accurate poses when subjects are clearly visible, but lacks complete temporal coverage.

We combine both of these scenarios into a single unified conditional diffusion model $p_\theta (S^i|C^i,\hat{S}^{i1},\ldots,\hat{S}^{iN})$ which is designed to integrate variable numbers of cameras and observations $\{\hat{S}^{ij}\}$ which are sparse or intermittent in time.

\subsection{Reconstruction via conditional diffusion}
We develop a diffusion-based model for $p_\theta (S^i \mid \{ \hat{S}^{ij} \},C^i)$ that conditions on the egocentric motion $C^i$ and exo-observations $\{ \hat{S}^{ij} \}$ which vary in number at each timestep.
\cref{fig:architecture} gives an overview of our model architecture for modeling this problem using 
conditional diffusion.

We use a Diffusion Transformer (DiT)-based conditional denoiser architecture (${\sim}54$M parameters) to estimate the state for the given target individual. At each step of the diffusion process, the diffusion model takes as input the current (noisy) estimate of the body trajectory  $S^i$ and conditioning observations $\{\hat{S}^{ij}\}$ and ego-motion trajectory $C^i$. We encode the current noisy body trajectory estimate as well as any exocentric observations using a common body encoder. Each observer's encoded sequence is processed by a temporal self-attention module shared across observers. Pooled observation encoding is then fused additively at multiple layers of the DiT. The ego trajectory $C^i$ is encoded separately and integrated by cross-attention at each layer of the DiT.

In order for the model to accommodate variable numbers of observations at each frame, we rely on attentional pooling to fuse them into a single summary feature.
On frames with no active observers, the fusion residual is gated to zero.
During training we perform augmentation by dropping out observations so the model learns to function well with sparse observations.

\subsection{Alternative approaches to fusing observations}

In addition to our full conditional model, we also explored several alternatives for handling conditioning on exo-observations but still using the same denoising transformer and camera trajectory conditioning architecture. 

\noindent\textbf{Imputation.} Since the exo-observations $\{ \hat{S}^{ij} \}$ live in the same space as the pose $S^i$ being estimated and are more accurate than pure ego-motion-based estimates when available, it is natural to consider a setting where observations are assumed to be the best estimates and the model only needs to impute the missing observations. One potential advantage of this approach is that we can train an ego-only model $p_\theta (S^i \mid C^i)$ without specifying an observation sparsity or noise distribution, and then adapt it at inference time depending on the available observations.
This can be implemented by running the forward diffusion process on the observations and overwriting the observed component during each step of reverse diffusion \cite{song2021scorebased}.  This has been applied to adapt a pretrained diffusion model for image inpainting tasks \cite{repaint} and human motion models for ``in-betweening'' specified key-frames~\cite{priormdm, condmdi}.

\noindent\textbf{Conditioning/imputation hybrid.}
CondMDI~\cite{condmdi} is a hybrid approach that directly overwrites the observed/constrained dimensions of the noisy sample trajectory at each step of the iterative denoising process (rather than using the forward model). However, it is trained with additional conditioning in the form of a binary mask that specifies which dimensions were observed.
This form of conditioning is effective at quickly guiding the diffusion process to the observed state but in our setting, performance is constrained by the noise floor of the observation model.  
We found empirically that performance could be improved upon by a heuristic of only overwriting with observations when the diffusion noise is large (e.g., $t>20$) and switching to the unconditional denoiser for the remaining iterations to allow the model to perform fine-grained refinement, further improving sample quality below the noise floor of the observations.

\noindent\textbf{Diffusion posterior sampling (DPS).}
DPS~\cite{dps} assumes a Gaussian noise model for observations which can be easily estimated from training examples. It then uses the corresponding conditional score function as ``regression guidance'' during diffusion. In this setting, the conditional component contributes significantly to sampling updates early during reverse-diffusion but the prior score function plays a bigger role in the vicinity of the final solution. This is less heuristic than imputation but still makes strong simplifying assumptions on the observation noise which instead can be learned directly in our fully conditional fusion approach.

\section{Experiments}
\label{sec:experiments}

\begin{table*}[t]
  \centering
  \caption{\small
  EgoHumans test ($n=1869$).}
  \vspace{-2mm}
  \label{tab:egohumans_full}
  \setlength{\tabcolsep}{2mm}
  \scalebox{0.95}{
  \begin{tabular}{lcccccc}
    \toprule
    Method & MPJPE$\downarrow$ & WA-MPJPE$\downarrow$ & PA-MPJPE$\downarrow$ & MPJAE$\downarrow$ & MPJVE$\downarrow$ & Accel$\downarrow$ \\
    \midrule
    Egocentric Motion                & 111.9          & 105.0          & 82.6          & 29.4          & 384.0          &  7760          \\
    Exocentric Video (CoMotion+SLAM) & 130.3          & 108.8          & 50.9          & 26.8          & 410.4          &  7909          \\
    Imputation (RePaint)             & 121.2          & 104.1          & 49.2          & 25.5          & 654.1          & 22371          \\
    Imputation (CondMDI)             &  97.6          &  86.2          & 56.9          & 22.3          & 367.0          &  7892          \\
    Posterior Sampling (DPS)         &  92.1          &  82.2          & 55.0          & 22.8          & \textbf{332.5} & \textbf{7212}  \\
    Ours                             & \textbf{90.7}  & \textbf{78.6}  & \textbf{45.9} & \textbf{19.7} & 357.0          &  7678          \\
    \bottomrule
  \end{tabular}
  }
\end{table*}

\begin{table*}[t]
  \centering
  \caption{\small
  Harmony4D-ego test ($n=496$).
  Single-observer only (0--1 observers per frame).}
  \vspace{-2mm}
  \label{tab:h4d_ego}
  \setlength{\tabcolsep}{2mm}
  \scalebox{0.95}{
  \begin{tabular}{lcccccc}
    \toprule
    Method & MPJPE$\downarrow$ & WA-MPJPE$\downarrow$ & PA-MPJPE$\downarrow$ & MPJAE$\downarrow$ & MPJVE$\downarrow$ & Accel$\downarrow$ \\
    \midrule
    Egocentric Motion                &  95.6          &  93.8          & 73.3          & 24.1          & 324.0          &  4917          \\
    Exocentric Video (CoMotion+SLAM) & 147.9          & 119.0          & 60.6          & 27.9          & 349.1          &  5010          \\
    Imputation (RePaint)             & 118.3          & 101.8          & 50.6          & 22.8          & 496.4          & 13935          \\
    Imputation (CondMDI)             &  79.4          &  74.1          & 51.2          & 18.5          & 298.4          &  4919          \\
    Posterior Sampling (DPS)         &  85.0          &  79.7          & 56.3          & 19.5          & 293.6          & \textbf{4492}  \\
    Ours                             & \textbf{73.4}  & \textbf{67.8}  & \textbf{40.9} & \textbf{15.9} & \textbf{284.4} &  4747          \\
    \bottomrule
  \end{tabular}
  }
\end{table*}

\begin{figure*}[tp]
\centering
\includegraphics[width=\textwidth]{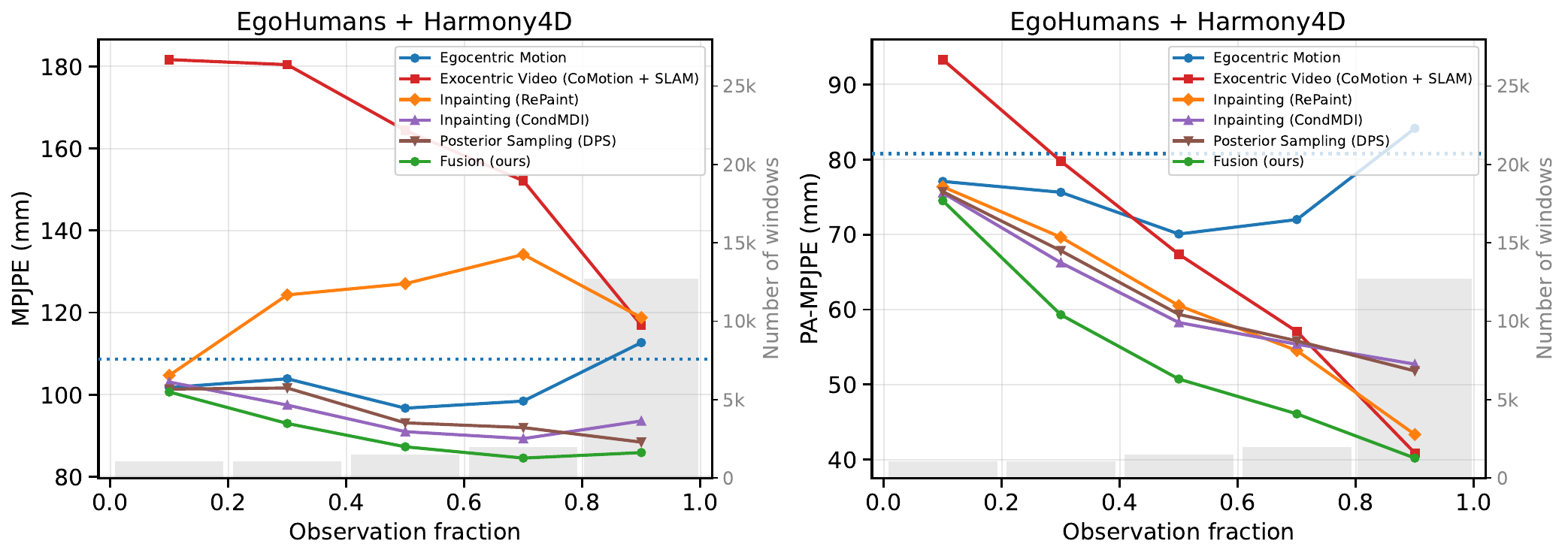}
\vspace{-6mm}
\caption{\small
Mean per-joint position error as a function of observation fraction (fraction of frames where at least one camera detects the subject). 
Gray bars show the number of evaluation windows at each observation fraction.
Egocentric motion estimation doesn't use observations and is roughly flat (dotted blue line shows the average error).  Exocentric video estimation steadily improves with observation fraction. In terms of absolute error (in world coordinates) ego motion and exo video perform similarly for high observation fraction. However, exo video provides better estimates of relative pose (Procrustes aligned or joint angles) at high obs fractions.
Observation replacement interpolates between these as fraction increases (performs like ego for low obs fraction, exo for high obs fraction). The learned conditional models (CondMDI, DPS, Fusion) all show benefits from taking observation noise into account and allow for ``synergistic fusion'', not just interpolation. Our learned conditional model achieves best performance across all obs fractions.
}
\label{fig:obs_frac}
\end{figure*}

We evaluate fusion of egocentric motion and exocentric observations in dynamic in-the-wild settings.

\subsection{Datasets}

\noindent\textbf{AMASS.}
We train on AMASS~\cite{amass}, a large-scale motion capture dataset providing diverse single-person SMPL-H motion sequences at 30 FPS.
Following EgoAllo~\cite{egoallo}, we use AMASS to train our conditional diffusion prior.
Since AMASS contains only motion capture data without real head-mounted cameras, we synthesize egocentric device coordinates from the body meshes, defining a mesh-derived Central Pupil Frame (CPF) at the midpoint between the eyes with orientation aligned to the head joint, as in EgoAllo.

\noindent\textbf{EgoHumans.}
We train and evaluate on EgoHumans~\cite{egohumans}, which uniquely provides 3D annotations for all participants in a fully egocentric multi-camera setup where each person wears a head-mounted Aria camera, enabling our study of mutual egocentric observations between wearers.
The dataset features dynamic, in-the-wild activities with 3--4 participants per scene.
For each scene, we use RGB video streams, camera trajectories from VIO, and pseudo-ground-truth SMPL body parameters obtained through multi-view triangulation and optimization.
The original 20 FPS capture is resampled to 30 FPS to match our model's training frame rate.

The Aria device defines the CPF as the reference frame for head tracking. EgoHumans provides RGB camera trajectories, which we convert to CPF poses using the device's constant RGB-to-CPF transform. To align the real Aria CPF with our mesh-derived CPF from AMASS training, we apply an averaged offset parameterized as a pitch rotation of 14.84$^\circ$ with translations of 1.2mm downward and 18mm backward, estimated across EgoHumans wearers.

\noindent\textbf{Harmony4D.}
Harmony4D~\cite{harmony4d} contains 208 sequences of close-contact two-person
interactions across six activity categories (hugging, grappling, sword,
ballroom, karate, MMA), totaling 115k frames at 30 FPS upsampled from the
native 20 FPS. Of these, 118 are egocentric, with both participants wearing
head-mounted Aria cameras. In both EgoHumans and Harmony4D, we follow the released train/test partition. For EgoHumans, the test activities (fencing, lego assembly, tagging) do not appear in the training split.

\subsection{Implementation details}

\noindent\textbf{Egocentric motion prior.}
Our egocentric motion prior (\cref{sec:method}) conditions on the
wearer's head trajectory alone. We train it from scratch on a mixture of AMASS,
EgoHumans, and Harmony4D for 60{,}000 steps at a global batch size of 2048 on
8$\times$NVIDIA RTX PRO 6000 GPUs. We initialize both our conditional fusion model and the
trained imputation baseline (CondMDI~\cite{condmdi}) from this prior, so that
comparisons isolate the conditioning mechanism rather than the prior itself.

\noindent\textbf{Conditional fine-tuning.}
Starting from the motion prior, we fine-tune our conditional fusion model at the
same batch size for 20{,}000 steps. For the first 5{,}000 steps, we freeze the
backbone and train only the new fusion modules on EgoHumans and Harmony4D; we
then unfreeze the full network and add AMASS to the data mix. Because AMASS
contains no visual observations, we synthesize them by perturbing ground-truth
body rotations with noise calibrated to match the magnitude of CoMotion's
per-joint rotation error on real data. The CondMDI baseline is fine-tuned from
the same prior with the same data and synthetic-noise scheme.

\noindent\textbf{Augmentation and inference.}
We apply left/right flip augmentation~\cite{egoallo} to all training data.
During fine-tuning we resample each window's observation fraction uniformly from
$(0,1)$~\cite{condmdi} and drop all of a window's observations with probability
$0.1$, exposing the model to coverage levels from fully unconditional to fully
observed. At inference, we run 30 DDIM steps over a $T{=}1000$ cosine noise
schedule with quadratic timestep spacing and deterministic sampling ($\eta{=}0$),
and evaluate EMA weights. Complete optimizer, schedule, and augmentation
settings are listed in the appendix.

\subsection{Metrics}

We evaluate accuracy in absolute positioning, relative body structure, and
motion quality.  All estimates and evaluations use a shared world coordinate
frame ($z$-up, floor at $z=0$). The rigid transform between the device
trajectory and the training CPF is estimated per person per sequence from
calibration data.

\noindent\textbf{Global accuracy.} MPJPE (mean per-joint position error in mm)
measures absolute joint accuracy in world coordinates.  We also consider the
``world aligned'' WA-MPJPE which rigidly aligns the whole trajectory. This
discounts constant coordinate-frame offsets while preserving trajectory-level
consistency.

\noindent\textbf{Relative pose accuracy.} PA-MPJPE applies Procrustes alignment
(rotation, translation, scale) before computing errors, isolating pose quality
independent of global positioning. MPJAE measures mean per-joint angle error in
degrees after alignment, ignoring global translations.

\noindent\textbf{Temporal consistency.} MPJVE (mean per-joint velocity error in
m/s) evaluates whether joints move with correct speed and direction. Accel
measures acceleration error (m/s$^2$), quantifying velocity discontinuities
that cause jitter.

\subsection{Fusion outperforms egocentric or exocentric baselines}

We evaluate our proposed Fusion model along with five baselines: the egocentric motion estimate, exocentric visual tracking with device alignment (CoMotion+SLAM), imputation (RePaint), replacement with explicit masking (CondMDI), and guidance with learned variance (DPS).
\cref{tab:egohumans_full,tab:h4d_ego} compare the evaluated methods on EgoHumans and Harmony4D. The two standalone baselines each capture a different aspect of body estimation: the egocentric motion prior provides full temporal coverage with calibrated global positioning (111.9mm MPJPE on EgoHumans) but poor relative body pose (82.6mm PA-MPJPE), while visual tracking with device alignment achieves much better relative pose (50.9mm PA-MPJPE) but worse global error (130.3mm MPJPE).%
The observation-conditioned diffusion approach combines both signals. It improves on both axes simultaneously, reaching 90.7mm MPJPE and 45.9mm PA-MPJPE on EgoHumans. None of the baselines dominates both metrics.

We further analyze performance on a pooled test set as a function of visibility. Observation fraction is the fraction of frames within each 128-frame window where at least one camera detects the subject evaluated over the pooled datasets.  Across observation fractions (\cref{fig:obs_frac}), the fusion is strictly additive: the conditioned model performs at least as well as the egocentric-only prior at every coverage level and improves further with additional observations, with no degradation at low coverage.  These improvements are consistent across both datasets, with Harmony4D (\cref{tab:h4d_ego}) showing the same pattern: 73.4mm MPJPE and 40.9mm PA-MPJPE.

\subsection{Observation fraction and noise explain source reliability}

\noindent\textbf{Observation noise.}
When a person is visible, CoMotion provides accurate relative body pose that is substantially better than what head motion alone can achieve (\cref{fig:comparison}). However, it has three limitations. First, coverage is incomplete: temporal coverage varies from 60\% (lego assembly, where participants focus on a shared object) to 89\% (fencing, with close face-to-face interaction) across activity categories in our evaluation dataset. Second, absolute positioning is unreliable due to monocular scale ambiguity in the per-frame 3D estimates. Third, the predicted body rotations are noisy, with a mean geodesic error of approximately $17^\circ$ per joint relative to ground truth, and the spine2 joint alone contributes disproportionately to this error.

\noindent\textbf{Analysis.}
We decompose the evaluation into 128-frame windows (${\sim}4.3$ seconds, matching the model's input size), sampled at a denser stride than the headline tables, and analyze error as a function of observation conditions.

The 18{,}339 resulting (subject, window) pairs span a wide range of observation conditions. Only 4.4\% of windows have near-zero exocentric observations (observation fraction $< 0.1$), while 59\% have coverage above 0.9; the median observation fraction is 0.97. This distribution is a property of the evaluation setting, not the method: it reflects the geometry of multi-person egocentric interaction, where mutual visibility depends on spatial arrangement, the camera's limited field of view (${\sim}110^\circ$ for Aria glasses), and activity structure. Because the distribution is skewed toward high coverage, the aggregate MPJPE is dominated by the system's behavior when observations are plentiful.

\noindent\textbf{Error and observation coverage.}
\cref{fig:obs_frac} shows mean MPJPE as a function of observation fraction. The egocentric motion prior varies only mildly from the average (dottedline: 109mm average)  across coverage levels and does not decrease, indicating that higher-coverage windows are not intrinsically easier. Visual tracking with device alignment (CoMotion+SLAM) degrades sharply as coverage drops, as missing frames are filled by SLERP interpolation across widening temporal gaps; its global error approaches the prior only at the highest coverage.

Notably, the conditional fusion model is able to improve even when all frames are observed. At 100\% coverage, exocentric pose 
estimates alone produce
107.2mm MPJPE and 37.5mm PA-MPJPE across all frames. This is due to the inherent noise in human pose estimation.
And the motion prior helps us bring the global error down to 88.2mm MPJPE (19mm improvement) while taking a 1.5mm hit on PA-MPJPE.
This differentiates our setting from the majority of in-betweening methods that focus on interpolating between accurate keyframes.

\noindent\textbf{Activity categories.}
Performance varies across activity categories, partly explained by differences in mutual visibility.
Observation fraction varies substantially across the test activities: fencing (89\%) and tagging (88\%),
where participants closely interact, have high observation rates, while lego assembly drops to 60\% as participants focus on a shared object
and frequently occlude one another. MPJPE follows this pattern: fencing achieves 88mm and tagging 89mm, while lego assembly is 100mm. The
pattern is clearer for the visual-tracking baseline (CoMotion+SLAM), which degrades much more sharply with reduced coverage (115mm $\to$ 132mm $\to$ 163mm), reflecting its direct dependence on per-frame visibility;
our fusion model relies on the egocentric prior to absorb the missing frames, which is why its
degradation across the same coverage range is roughly a quarter (12mm vs 48mm).

\subsection{Learned conditioning outperforms imputation and guidance}

\cref{tab:egohumans_full,tab:h4d_ego} compare our learned conditioning against three baselines that share the same diffusion backbone but differ in how observations enter the denoising process: hard imputation (RePaint), mask-conditioned imputation (CondMDI), and posterior sampling (DPS). Our model improves over the best of these on accuracy metrics (MPJPE, PA-MPJPE, MPJAE) on both datasets. Temporal smoothness is comparable to DPS, which preserves the prior's smoothness via gradient guidance without constraining frames.

The observation values come from CoMotion's SMPL parameter estimates. Because SMPL local joint rotations (joints 1--21) are frame-invariant and identical regardless of the world or camera coordinate system, CoMotion's predictions can be used directly as replacement values without coordinate transformation. For frames with a valid observation, the replacement-based baselines (RePaint, CondMDI) overwrite the betas (10 dims) and body rotations (126 dims) with these values. Replacement is all-or-nothing at the frame level: frames without a valid observation receive no replacement and are generated entirely by the diffusion prior.

Conditioning on only the highest-confidence exocentric observation at each timestep performs slightly worse (91.8 MPJPE / 46.3 PA-MPJPE on EgoHumans) than attentional pooling over all available observations (90.7 / 45.9).%

\section{Discussion and Conclusion}
\label{sec:discussion}
We present an approach for multi-person pose estimation from egocentric cameras, integrating visual tracking with generative motion modeling. Each participant's head-mounted device provides a head-motion prior, and body estimates from other participants' cameras complement this with visually grounded pose. Anchoring these exocentric estimates to the subject's own device trajectory via SLAM alignment improves their global positioning. Because the diffusion model conditions on SMPL body parameters from the tracker, the framework is not tied to any particular tracking or prior model. As more capable models become available, they can be used directly.

This modular structure also has practical implications: per-camera detection can be run on-device, and only low-bandwidth pose summaries need to be exchanged between participants, in contrast to end-to-end approaches that would need access to all raw video.

\noindent\textbf{Broader impacts.}
Egocentric data carries privacy concerns: wearable cameras capture people and environments beyond the wearer. We focus on collaborative sharing among individuals in a shared environment, which differs from unilateral exocentric surveillance and assumes informed consent and trust among all participants. The underlying building blocks are nevertheless shared, and responsible deployment requires explicit agreement from everyone being tracked.

\noindent\textbf{Limitations and future work.}
Acquiring high-quality data for training and evaluation remains a challenge. Gold-standard motion capture data (e.g., AMASS) differs in distribution from real-world multi-person activities. In our study, we evaluate on smaller but more realistic datasets (EgoHumans and Harmony4D) but ground truth is imperfect and the amount of data is limited. How well the approach generalizes to other settings, with different activity types, participant counts, or device configurations, remains to be established.
In the current formulation, people in the scene are coupled through the observer-subject relationship: when one person's camera captures another, those observations inform the subject's pose estimate. A natural extension is to also model physical and social interactions between the estimated trajectories themselves. People collide, coordinate, and react to one another: a basketball player's motion depends on the defender's position, fencers make contact, and conversational partners adjust their body language in response to each other. Incorporating these dependencies, through inter-person attention or collision constraints within the diffusion and guidance processes, could improve plausibility, especially in close-interaction settings.

{
  \small
  \bibliographystyle{ieeenat_fullname}
  \bibliography{main}

\begin{thebibliography}{34}
\providecommand{\natexlab}[1]{#1}
\providecommand{\url}[1]{\texttt{#1}}
\expandafter\ifx\csname urlstyle\endcsname\relax
  \providecommand{\doi}[1]{doi: #1}\else
  \providecommand{\doi}{doi: \begingroup \urlstyle{rm}\Url}\fi

\bibitem[Chen et~al.(2023)Chen, Zhang, Li, Pang, Xia, and Liu]{Chen_2023_ICCV}
Ling-Hao Chen, Jiawei Zhang, Yewen Li, Yiren Pang, Xiaobo Xia, and Tongliang
  Liu.
\newblock {HumanMAC}: Masked motion completion for human motion prediction.
\newblock In \emph{ICCV}, pages 9544--9555, 2023.

\bibitem[Chen et~al.(2026)Chen, Chen, Xue, Chen, Xiu, and
  Pons-Moll]{chen2025human3r}
Yue Chen, Xingyu Chen, Yuxuan Xue, Anpei Chen, Yuliang Xiu, and Gerard
  Pons-Moll.
\newblock {Human3R}: Everyone everywhere all at once.
\newblock In \emph{ICLR}, 2026.

\bibitem[Chung et~al.(2023)Chung, Kim, McCann, Klasky, and Ye]{dps}
Hyungjin Chung, Jeongsol Kim, Michael~T. McCann, Marc~L. Klasky, and Jong~Chul
  Ye.
\newblock Diffusion posterior sampling for general noisy inverse problems.
\newblock In \emph{ICLR}, 2023.

\bibitem[Cohan et~al.(2024)Cohan, Tevet, Reda, Peng, and van~de Panne]{condmdi}
Setareh Cohan, Guy Tevet, Daniele Reda, Xue~Bin Peng, and Michiel van~de Panne.
\newblock Flexible motion in-betweening with diffusion models.
\newblock In \emph{SIGGRAPH}, 2024.

\bibitem[Jiang et~al.(2024)Jiang, Streli, Meier, and Holz]{egoposer}
Jiaxi Jiang, Paul Streli, Manuel Meier, and Christian Holz.
\newblock {EgoPoser}: Robust real-time egocentric pose estimation from sparse
  and intermittent observations everywhere.
\newblock In \emph{ECCV}, pages 277--294, 2024.

\bibitem[Kanazawa et~al.(2018)Kanazawa, Black, Jacobs, and Malik]{hmr}
Angjoo Kanazawa, Michael~J. Black, David~W. Jacobs, and Jitendra Malik.
\newblock End-to-end recovery of human shape and pose.
\newblock In \emph{CVPR}, pages 7122--7131, 2018.

\bibitem[Khirodkar et~al.(2023)Khirodkar, Bansal, Ma, Newcombe, Vo, and
  Kitani]{egohumans}
Rawal Khirodkar, Aayush Bansal, Lingni Ma, Richard Newcombe, Minh Vo, and Kris
  Kitani.
\newblock {EgoHumans}: An egocentric {3D} multi-human benchmark.
\newblock In \emph{ICCV}, pages 19807--19819, 2023.

\bibitem[Khirodkar et~al.(2024)Khirodkar, Song, Cao, Luo, and
  Kitani]{harmony4d}
Rawal Khirodkar, Jyun-Ting Song, Jinkun Cao, Zhengyi Luo, and Kris Kitani.
\newblock {Harmony4D}: A video dataset for in-the-wild close human
  interactions.
\newblock In \emph{NeurIPS}, 2024.

\bibitem[Kocabas et~al.(2020)Kocabas, Athanasiou, and Black]{vibe}
Muhammed Kocabas, Nikos Athanasiou, and Michael~J. Black.
\newblock {VIBE}: Video inference for human body pose and shape estimation.
\newblock In \emph{CVPR}, pages 5253--5263, 2020.

\bibitem[Li et~al.(2025)Li, Cao, Zhang, Rempe, Kautz, Iqbal, and
  Yuan]{li2025genmo}
Jiefeng Li, Jinkun Cao, Haotian Zhang, Davis Rempe, Jan Kautz, Umar Iqbal, and
  Ye Yuan.
\newblock {GENMO}: A {GENeralist} model for human {MOtion}.
\newblock In \emph{ICCV}, pages 11766--11776, 2025.

\bibitem[Liu et~al.(2026)Liu, Lin, Wu, and Zhou]{liu2025josh}
Zhizheng Liu, Joe Lin, Wayne Wu, and Bolei Zhou.
\newblock Joint optimization for {4D} human-scene reconstruction in the wild.
\newblock In \emph{ICLR}, 2026.

\bibitem[Loper et~al.(2015)Loper, Mahmood, Romero, Pons-Moll, and Black]{smpl}
Matthew Loper, Naureen Mahmood, Javier Romero, Gerard Pons-Moll, and Michael~J.
  Black.
\newblock {SMPL}: A skinned multi-person linear model.
\newblock \emph{ACM TOG}, 34\penalty0 (6):\penalty0 248:1--248:16, 2015.
\newblock Proc. SIGGRAPH Asia.

\bibitem[Lugmayr et~al.(2022)Lugmayr, Danelljan, Romero, Yu, Timofte, and
  Van~Gool]{repaint}
Andreas Lugmayr, Martin Danelljan, Andr{\'e}s Romero, Fisher Yu, Radu Timofte,
  and Luc Van~Gool.
\newblock {RePaint}: Inpainting using denoising diffusion probabilistic models.
\newblock In \emph{CVPR}, pages 11461--11471, 2022.

\bibitem[Mahmood et~al.(2019)Mahmood, Ghorbani, Troje, Pons-Moll, and
  Black]{amass}
Naureen Mahmood, Nima Ghorbani, Nikolaus~F. Troje, Gerard Pons-Moll, and
  Michael~J. Black.
\newblock {AMASS}: Archive of motion capture as surface shapes.
\newblock In \emph{ICCV}, pages 5442--5451, 2019.

\bibitem[Newell et~al.(2025)Newell, Hu, Lipson, Richter, and Koltun]{comotion}
Alejandro Newell, Peiyun Hu, Lahav Lipson, Stephan~R. Richter, and Vladlen
  Koltun.
\newblock {CoMotion}: Concurrent multi-person {3D} motion.
\newblock In \emph{ICLR}, 2025.

\bibitem[Ng et~al.(2020)Ng, Xiang, Joo, and Grauman]{ng_you2me_2020}
Evonne Ng, Donglai Xiang, Hanbyul Joo, and Kristen Grauman.
\newblock {You2Me}: Inferring body pose in egocentric video via first and
  second person interactions.
\newblock In \emph{CVPR}, pages 9890--9900, 2020.

\bibitem[Rempe et~al.(2021)Rempe, Birdal, Hertzmann, Yang, Sridhar, and
  Guibas]{Rempe_2021_ICCV}
Davis Rempe, Tolga Birdal, Aaron Hertzmann, Jimei Yang, Srinath Sridhar, and
  Leonidas~J. Guibas.
\newblock {HuMoR}: {3D} human motion model for robust pose estimation.
\newblock In \emph{ICCV}, pages 11488--11499, 2021.

\bibitem[Romero et~al.(2017)Romero, Tzionas, and Black]{smplh}
Javier Romero, Dimitrios Tzionas, and Michael~J. Black.
\newblock Embodied hands: Modeling and capturing hands and bodies together.
\newblock \emph{ACM TOG}, 36\penalty0 (6):\penalty0 245:1--245:17, 2017.
\newblock Proc. SIGGRAPH Asia.

\bibitem[Saadatnejad et~al.(2023)Saadatnejad, Rasekh, Mofayezi, Medghalchi,
  Rajabzadeh, Mordan, and Alahi]{saadatnejad_diffusion_pose}
Saeed Saadatnejad, Ali Rasekh, Mohammadreza Mofayezi, Yasamin Medghalchi, Sara
  Rajabzadeh, Taylor Mordan, and Alexandre Alahi.
\newblock A generic diffusion-based approach for {3D} human pose prediction in
  the wild.
\newblock In \emph{ICRA}, 2023.

\bibitem[Shafir et~al.(2024)Shafir, Tevet, Kapon, and Bermano]{priormdm}
Yonatan Shafir, Guy Tevet, Roy Kapon, and Amit~H. Bermano.
\newblock Human motion diffusion as a generative prior.
\newblock In \emph{ICLR}, 2024.

\bibitem[Shen et~al.(2024)Shen, Pi, Xia, Cen, Peng, Hu, Bao, Hu, and
  Zhou]{shen2024gvhmr}
Zehong Shen, Huaijin Pi, Yan Xia, Zhi Cen, Sida Peng, Zechen Hu, Hujun Bao,
  Ruizhen Hu, and Xiaowei Zhou.
\newblock World-grounded human motion recovery via gravity-view coordinates.
\newblock In \emph{SIGGRAPH Asia}, 2024.

\bibitem[Song et~al.(2021)Song, Sohl-Dickstein, Kingma, Kumar, Ermon, and
  Poole]{song2021scorebased}
Yang Song, Jascha Sohl-Dickstein, Diederik~P. Kingma, Abhishek Kumar, Stefano
  Ermon, and Ben Poole.
\newblock Score-based generative modeling through stochastic differential
  equations.
\newblock In \emph{ICLR}, 2021.

\bibitem[Sun et~al.(2023)Sun, Bao, Liu, Mei, and Black]{Sun_2023_CVPR}
Yu Sun, Qian Bao, Wu Liu, Tao Mei, and Michael~J. Black.
\newblock {TRACE}: {5D} temporal regression of avatars with dynamic cameras in
  {3D} environments.
\newblock In \emph{CVPR}, pages 8856--8866, 2023.

\bibitem[Tevet et~al.(2023)Tevet, Raab, Gordon, Shafir, Cohen-Or, and
  Bermano]{mdm}
Guy Tevet, Sigal Raab, Brian Gordon, Yonatan Shafir, Daniel Cohen-Or, and
  Amit~H. Bermano.
\newblock Human motion diffusion model.
\newblock In \emph{ICLR}, 2023.

\bibitem[von Marcard et~al.(2018)von Marcard, Henschel, Black, Rosenhahn, and
  Pons-Moll]{vonMarcard2018}
Timo von Marcard, Roberto Henschel, Michael~J. Black, Bodo Rosenhahn, and
  Gerard Pons-Moll.
\newblock Recovering accurate {3D} human pose in the wild using {IMUs} and a
  moving camera.
\newblock In \emph{ECCV}, pages 601--617, 2018.

\bibitem[Wang et~al.(2024)Wang, Wang, Liu, and Daniilidis]{wang2024tram}
Yufu Wang, Ziyun Wang, Lingjie Liu, and Kostas Daniilidis.
\newblock {TRAM}: Global trajectory and motion of {3D} humans from in-the-wild
  videos.
\newblock In \emph{ECCV}, pages 467--487, 2024.

\bibitem[Wang et~al.(2026)Wang, Ng, Shin, Khirodkar, Dong, Su, Park, Kitani,
  Richard, Prada, and Zollh{\"o}fer]{wang2026duomo}
Yufu Wang, Evonne Ng, Soyong Shin, Rawal Khirodkar, Yuan Dong, Zhaoen Su,
  Jinhyung Park, Kris Kitani, Alexander Richard, Fabian Prada, and Michael
  Zollh{\"o}fer.
\newblock {DuoMo}: Dual motion diffusion for world-space human reconstruction.
\newblock In \emph{CVPR}, 2026.

\bibitem[Xie et~al.(2024)Xie, Jampani, Zhong, Sun, and Jiang]{omnicontrol}
Yiming Xie, Varun Jampani, Lei Zhong, Deqing Sun, and Huaizu Jiang.
\newblock {OmniControl}: Control any joint at any time for human motion
  generation.
\newblock In \emph{ICLR}, 2024.

\bibitem[Yang et~al.(2026)Yang, Kukreja, Pinkus, Sagar, Fan, Park, Shin, Cao,
  Liu, Ugrinovic, Feiszli, Malik, Doll{\'a}r, and Kitani]{yang2026sam3dbody}
Xitong Yang, Devansh Kukreja, Don Pinkus, Anushka Sagar, Taosha Fan, Jinhyung
  Park, Soyong Shin, Jinkun Cao, Jiawei Liu, Nicolas Ugrinovic, Matt Feiszli,
  Jitendra Malik, Piotr Doll{\'a}r, and Kris Kitani.
\newblock {SAM 3D Body}: Robust full-body human mesh recovery.
\newblock \emph{arXiv preprint arXiv:2602.15989}, 2026.

\bibitem[Ye et~al.(2023)Ye, Pavlakos, Malik, and Kanazawa]{ye2023decoupling}
Vickie Ye, Georgios Pavlakos, Jitendra Malik, and Angjoo Kanazawa.
\newblock Decoupling human and camera motion from videos in the wild.
\newblock In \emph{CVPR}, pages 21222--21232, 2023.

\bibitem[Yi et~al.(2025)Yi, Ye, Zheng, Li, M{\"u}ller, Pavlakos, Ma, Malik, and
  Kanazawa]{egoallo}
Brent Yi, Vickie Ye, Maya Zheng, Yunqi Li, Lea M{\"u}ller, Georgios Pavlakos,
  Yi Ma, Jitendra Malik, and Angjoo Kanazawa.
\newblock Estimating body and hand motion in an ego-sensed world.
\newblock In \emph{CVPR}, pages 7072--7084, 2025.

\bibitem[Yuan et~al.(2022)Yuan, Iqbal, Molchanov, Kitani, and Kautz]{glamr}
Ye Yuan, Umar Iqbal, Pavlo Molchanov, Kris Kitani, and Jan Kautz.
\newblock {GLAMR}: Global occlusion-aware human mesh recovery with dynamic
  cameras.
\newblock In \emph{CVPR}, pages 11038--11049, 2022.

\bibitem[Zhang et~al.(2022)Zhang, Ma, Zhang, Qian, Kwon, Pollefeys, Bogo, and
  Tang]{Zhang:ECCV:2022}
Siwei Zhang, Qianli Ma, Yan Zhang, Zhiyin Qian, Taein Kwon, Marc Pollefeys,
  Federica Bogo, and Siyu Tang.
\newblock {EgoBody}: Human body shape and motion of interacting people from
  head-mounted devices.
\newblock In \emph{ECCV}, pages 180--200, 2022.

\bibitem[Zhang et~al.(2024)Zhang, Bhatnagar, Xu, Winkler, Kadlecek, Tang, and
  Bogo]{rohm}
Siwei Zhang, Bharat~Lal Bhatnagar, Yuanlu Xu, Alexander Winkler, Petr Kadlecek,
  Siyu Tang, and Federica Bogo.
\newblock {RoHM}: Robust human motion reconstruction via diffusion.
\newblock In \emph{CVPR}, pages 14606--14617, 2024.

\end{thebibliography}
}

\iftoggle{cvprfinal}{%
  \newpage
  \appendix

\section{Appendix}

\subsection{Dataset pre-processing}
During training on AMASS (which has no wearable devices), the CPF is computed from the ground-truth body mesh following EgoAllo's derivation~\cite{egoallo}. At inference, the Aria device tracks the head trajectory using visual-inertial sensors, which does not correspond exactly to this virtual, mesh-derived CPF, due to variation in head shape and device placement. This difference is captured by a per-person rigid-transform prior estimated from calibration data. The prior assumes left-right symmetry and is held fixed across each sequence.

\subsection{Training and inference details}
The motion prior trains at a global batch size of 2048 (256 per GPU $\times$ 8), using Muon (learning rate $0.02$, weight decay $10^{-4}$) on the two-dimensional weight matrices and AdamW ($3{\times}10^{-4}$, weight decay $10^{-4}$) on the remaining parameters, with a 2{,}000-step linear warmup and cosine decay. Fine-tuning uses AdamW throughout: our conditional model at $3{\times}10^{-4}$ (1{,}000-step warmup for the frozen-backbone phase; 2{,}000-step warmup and cosine decay after unfreezing), and the CondMDI baseline at $10^{-4}$ in a single phase. Synthetic observations for AMASS perturb each ground-truth joint rotation by an isotropic Gaussian in the $\mathfrak{so}(3)$ tangent space with $\sigma{=}10^\circ$ per axis, giving a mean geodesic perturbation of ${\approx}16^\circ$, comparable to CoMotion's measured $17.1^\circ$ mean per-joint error; shape parameters are left clean, and synthetic observations carry neutral confidence and relative-pose metadata. Flip augmentation applies with $p{=}0.5$. During fine-tuning, the wearer's CPF conditioning is dropped per example with $p{=}0.05$; this maintains an unconditional branch, but no classifier-free guidance is used at inference. Evaluation uses EMA weights: decay $0.999$ for the fine-tuned models (ours, CondMDI) and $0.9999$ for methods running the frozen prior (motion prior, RePaint, DPS). All metrics are computed over independent 128-frame windows (stride 32), each denoised in a single pass with no cross-window stitching.

\subsection{Observation noise characteristics} Analysis of 180,931 observed frames reveals that CoMotion's body rotation predictions have a mean geodesic error of $17.1^\circ$ per joint. Approximately 49\% of this error is systematic bias (constant offset within temporal windows), and the noise is temporally correlated (lag-1 autocorrelation of 0.826). The high autocorrelation means that observation errors persist across frames rather than averaging out. Simple temporal averaging of consecutive observations would reduce i.i.d.\ noise but has limited effect on correlated errors. The spine2 joint dominates both rotation error ($24.6^\circ$) and its downstream effect on joint positions (43.8mm via forward kinematics), suggests that targeted improvement of trunk estimation in monocular trackers would have disproportionate impact on downstream fusion.

\subsection{Justification of variables for conditioning}
Detection confidence from the visual tracker correlates with prediction accuracy at the frame level (Spearman $\rho = -0.434$ between detection confidence and per-frame MPJPE).

\subsection{Motion prior backbone vs EgoAllo}
Our model without exocentric observations is functionally equivalent to the one
proposed in EgoAllo. We carried out a comparison using their training/test
split on AMASS alone and found that our improved DiT-based backbone model performs
significantly better than their released pretrained model. Performance improved
approximately 24mm MPJPE and 18mm PA-MPJPE on the AMASS test split
(\cref{tab:egoallo_backbone}).

\begin{table}[ht]
\centering
\caption{Comparison of our motion prior backbone with EgoAllo on the AMASS test split. All values in mm.}
\label{tab:egoallo_backbone}
\setlength{\tabcolsep}{12pt}
\scalebox{0.75}{
\begin{tabular}{lcc}
\toprule
Method & MPJPE $\downarrow$ & PA-MPJPE $\downarrow$ \\
\midrule
EgoAllo~\cite{egoallo} & 119.7 & 101.1 \\
Ours & \textbf{95.3} & \textbf{82.9} \\
\bottomrule
\end{tabular}
}
\end{table}

}{}

\end{document}